\documentclass{INTERSPEECH2023}

\interspeechcameraready

\usepackage{latexsym}
\usepackage{amsmath,amssymb}
\usepackage{mathtools}
\usepackage{multirow}
\usepackage[normalem]{ulem}
\useunder{\uline}{\uln}{}
\usepackage{graphicx}
\usepackage{boldline,multirow}
\usepackage{colortbl}
\definecolor{mygreen}{rgb}{0.0, 0.44, 0.0}
\usepackage{makecell}
\usepackage{boldline,multirow}
\usepackage{colortbl}
\usepackage{makecell}
\usepackage[colorinlistoftodos]{todonotes}
\usepackage{soul} 
\usepackage{balance}
\usepackage{subfig}
\newcommand{\asot}{CASSIE}
\newcommand{\ase}{AS2}
\usepackage{lipsum}

\title{Question-Context Alignment and Answer-Context Dependencies for \\Effective Answer Sentence Selection}
\name{Minh Van Nguyen$^{1*}$\thanks{* This work was completed while the author was an intern at Amazon Alexa AI.}, Kishan KC$^2$, Toan Nguyen$^2$, Thien Nguyen$^1$, Ankit Chadha$^2$, and Thuy Vu$^{2**}$\thanks{** Corresponding Author. Email: thuyvu@amazon.com}}
\address{
  $^1$Department of Computer Science, University of Oregon, OR, USA\\
  $^2$Amazon Alexa AI, CA, USA}
\email{\{minhnv,thien\}@cs.uoregon.edu\\\{ckshan,amztoan,ankitrc,thuyvu\}@amazon.com}

\begin{document}

\maketitle
 
\begin{abstract}
Answer sentence selection (AS2) in open-domain question answering finds answer for a question by ranking candidate sentences extracted from web documents. Recent work exploits answer context, i.e., sentences around a candidate, by incorporating them as additional input string to the Transformer models to improve the correctness scoring. In this paper, we propose to improve the candidate scoring by explicitly incorporating the dependencies between question-context and answer-context into the final representation of a candidate. Specifically, we use Optimal Transport to compute the question-based dependencies among sentences in the passage where the answer is extracted from. We then represent these dependencies as edges in a graph and use Graph Convolutional Network to derive the representation of a candidate, a node in the graph. Our proposed model achieves significant improvements on popular AS2 benchmarks, i.e., WikiQA and WDRASS, obtaining new state-of-the-art on all benchmarks.
\color{black}
\end{abstract}

\section{Introduction}

Voice-based virtual assistants powered by open-domain question answering (ODQA)~\cite{voorhees-1999-trec,wang-etal-2007-jeopardy} have gained significant commercial market in recent years, e.g., Google Assistant, Siri, or Alexa, thanks to the progress in question answering using large pre-trained language models (LLMs) such as BERT~\cite{devlin-etal-2019-bert}, RoBERTa~\cite{liu2019roberta}, and GPT-3~\cite{brown2020language}. 
Recent works address the task via generative models \cite{hsu-etal-2021-answer,nakano2021webgpt}. However, a well-known issue that has been shown to occur with the generative models is hallucination~\cite{maynez-etal-2020-faithfulness,roller-etal-2021-recipes,shuster-etal-2021-retrieval-augmentation}, where the models generate statements that are plausible looking but \emph{factually} incorrect.
Additionally, if the answers are composed by a pretrained LLM without external knowledge, the information contained in the answers might be outdated and no longer valid, e.g., the answer for the question \textit{``Which country is the reigning World Cup champion?''} will change through time. To avoid such problems, this work follows a typical ODQA pipeline involving two main stages: web retrieval and answer sentence selection ({\ase}); the latter selects the most relevant \emph{answer sentence} from the retrieved documents to return to the user.
This stage is typically implemented as a point-wise model that scores each sentence individually either without any additional information~\cite{wang-etal-2007-jeopardy,Garg_2020}, jointly with other top candidates~\cite{zhang-etal-2021-joint}, or with other contextual information~\cite{lauriola2021answer,han-etal-2021-modeling}. We focus on the latter in this paper.


Previous work on AS2 exploits contextual information for better performance, however, is limited to the concatenation approach. In particular, Lauriola et al.~\cite{lauriola2021answer} concatenated contextual sentences, i.e., previous (\texttt{prev}) and next sentences (\texttt{next}), with an answer candidate (\texttt{cand}) for a given question (\texttt{q}) as an input sequence to a Transformer architecture, i.e., using the structure ``\texttt{[CLS] q [SEP] prev [SEP] cand [SEP] next [EOS]}'' instead of ``\texttt{[CLS] q [SEP] cand [EOS]}.''
Han et al.~\cite{han-etal-2021-modeling}, similarly, uses a concatenation approach for contextual sentences and document titles.
While both models outperform the vanilla baseline, i.e., using only \texttt{cand}, we hypothesize that the improvement results from using a longer sequence for higher predictive power.

From a modeling viewpoint, the concatenation approach is effective largely due to the self-attention mechanism in the Transformer architecture~\cite{NIPS2017_3f5ee243}.
In particular, the mechanism allows weighing the relevance of each token with other tokens from the input, i.e., question, candidate, and contextual sentences. The approach, however, could be sub-optimal for AS2 for two reasons.
First, the concatenation approach fails to ignore irrelevant information in the context sentences, which could introduce noise to the prediction of the model.
Second, the approach fails to explicitly capture the alignment between the question and the answer/context sentences, which can be done to better reveal the correctness for the answer candidate. Table~\ref{tbl:intro-example} shows an example where a correct answer can be optimally selected by capturing the relevance between tokens from the question $q$ and sentences in a paragraph $p$ containing an answer candidate $a$.

\begin{table}[t]
\centering
\resizebox{\linewidth}{!}{%
\begin{tabular}{|lp{7.3cm}|}
\hline
$q$: & {What {\bf award} did {\bf Lionel Messi} win after the {\bf World Cup}?}\\
{} & {} \\
$a$: &\vspace{-0.8em}{\bf \color{mygreen} Lionel Messi has been crowned The Best FIFA Men's Player for the second time.}\\
{} & {} \\
$p_{a}$: &\vspace{-0.8em} {\texttt{[prev]}} The Best FIFA Football Awards annually honour the most outstanding members of the world's most popular sport. {\texttt{[cand]}} {\color{mygreen} {\bf Lionel Messi} has been crowned The Best FIFA Men's Player for the second time.} {\texttt{[next]}} The {\bf award}, which is voted for by national team coaches and captains, journalists and also fans, recognises a year in which the former Barcelona star crowned his glorious career by leading Argentina to victory at the {\bf World Cup}.\\
\hline
\end{tabular}
}
\caption{The example shows contextual sentences, i.e., \texttt{prev} and \texttt{next} in a passage $p$ for an answer $a$. Their relation with the question $q$ is critical to select  the {\bf \color{mygreen} correct answer} $a$.{\vspace{-3em}}}
\label{tbl:intro-example}
\end{table}

We address these shortcomings in this paper.
First, we propose to align the tokens of the question and the paragraph's sentences to enhance the relevance computing.
Specifically, we propose to employ Optimal Transport (OT) \cite{monge1781memoire,cuturi2013sinkhorn} to solve the question-context alignment problem.
Given two probability distributions over two point sets and a cost function that measures the distance between any two points, the goal of OT is to find a mapping that moves probability from one distribution to another such that the total cost of transporting the probability mass between two point sets is minimized.
We consider the question and a paragraph sentence (i.e., the answer candidate or context sentence) as two point sets, each word being a point.
To measure the distance between two points, we employ the Euclidean distance between their semantic representations, which can be obtained from a pre-trained language model (PLM), e.g., RoBERTa \cite{liu2019roberta}.
A probability distribution is also defined over each point set via the frequencies of the words in training data. Intuitively, the optimal alignment between the question and the candidate sentence maps the words that are both statistically and semantically similar between the two sentences. In the end, the relevant context formed by the set of the candidate words aligned with the question is utilized to compute the representation for the paragraph sentence.

Second, we address the dependencies among sentences in the paragraph, i.e., the answer candidate and its contextual sentences, to further bolster the modeling of the answer candidate representation.
In particular, we consider the paragraph sentences as nodes in a fully-connected graph and aim to learn a dependency weight between nodes.
To compute the dependency weight for two sentences, we propose to employ their semantic representations and transportation costs with respect to the question, which are already obtained from the question-context alignment step.
A feed-forward network with a sigmoid output function is then used to consume such information to estimate the dependency weight.
Afterwards, the dependency weights are utilized to enhance the representations for the sentences via a Graph Convolutional Network (GCN) \cite{kipf:17}. The output representation from the GCN for the answer candidate sentence can be directly sent to a binary classifier to obtain its correctness probability score. 
We finally exploit the mutual information (MI) between the GCN representations of the sentences to further encourage 
information sharing between them, i.e., by maximizing and minimizing the mutual information between the sentence pairs.
To this end, we treat the GCN representations of the sentences as continuous random variables.
The MI between the variables can then be optimized via the mutual information neural estimation (MINE) method \cite{belghazi2018mutual,hjelm2018learning}, which approximately estimates the lower bound of the MI via the binary cross entropy of a variable discriminator for optimization.

To demonstrate the effectiveness of our proposed model for AS2, we conduct experiments on a widely-used AS2 benchmark, i.e., WikiQA~\cite{yang-etal-2015-wikiqa}, used in previous work~\cite{lauriola2021answer} and a recent large-scale dataset WDRASS~\cite{zhang2022wdrass}.
Experimental results across the datasets show that our model achieves significant improvements compared to the previous work, obtaining new state-of-the-art performance for AS2.

\begin{figure}[ht]
    \centering
    \includegraphics[scale=0.33]{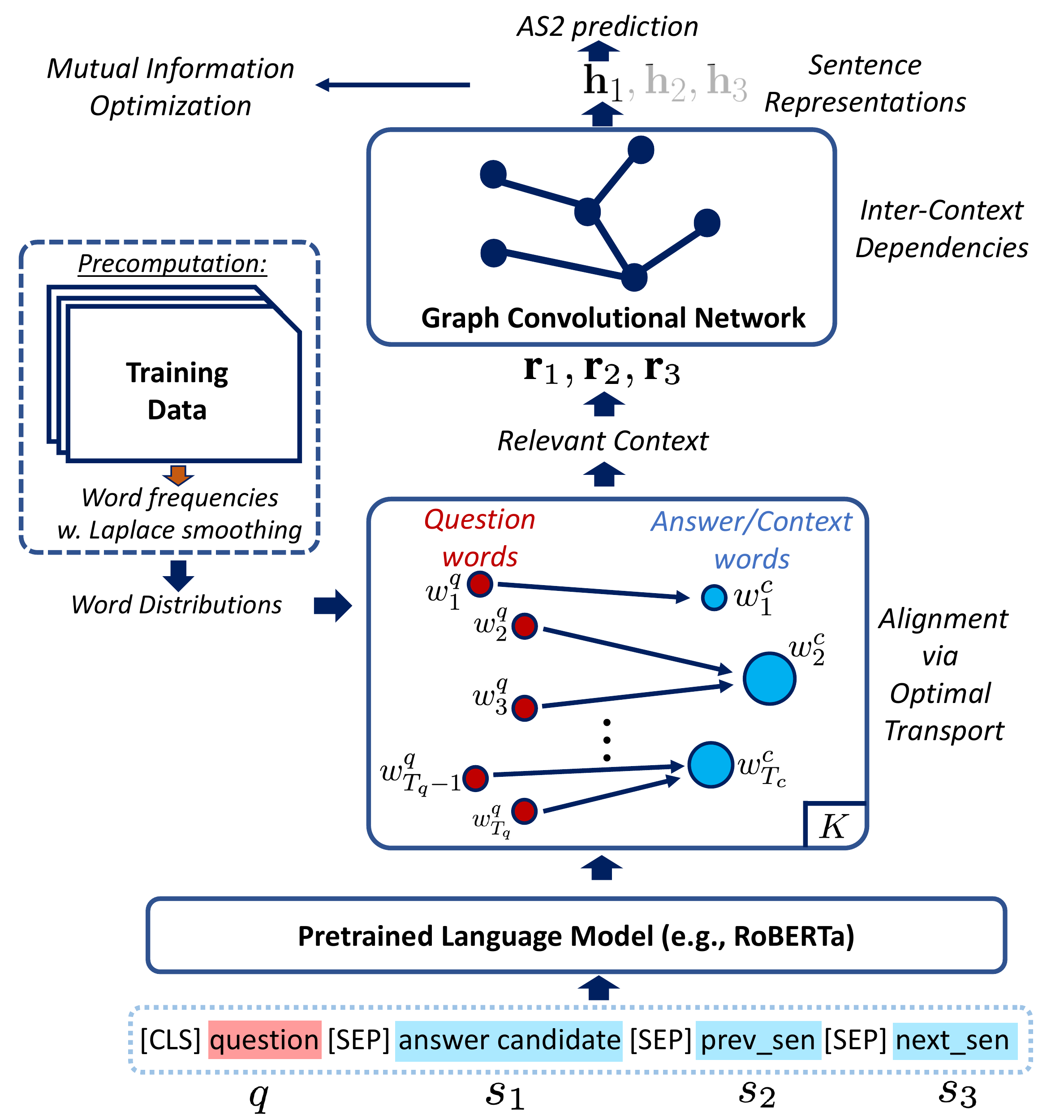}
    \caption{Illustration for our proposed model.}
    \label{fig:system-illustration}
\end{figure}

\color{black}

\section{Related work}

AS2 is an important task in ODQA and is often solved by point-wise methods that rank answer sentences extracted from retrieved Web documents \cite{severyn2015learning,shen-etal-2017-inter,yoon2019compare,Garg_2020}. 
Contextual information, e.g., neighboring sentences or document title, was recently incorporated to improve AS2~\cite{lauriola2021answer,han-etal-2021-modeling}.
Previous work, however, exploited the additional context by string concatenating~\cite{lauriola2021answer}, i.e., appending the input sequence.
By contrast, we propose to explicitly model the dependencies between the answer sentence and its context. 
We propose to align words between the question and the answer/context sentences via Optimal Transport (OT), a popular alignment method used in machine translation~\cite{dou-neubig-2021-word}, event argument extraction \cite{pouran-ben-veyseh-etal-2022-document}, and event coreference resolution \cite{phung-etal-2021-learning}. Our work is the first to apply OT to perform the question-context alignment for AS2.
 
\section{Architecture}

Given a question $q$ and a set of $N$ answer sentence candidates $C = \{c_1, c_2, \ldots, c_N \}$, the task of AS2 seeks to find correct answers $A \subset C$ via learning a reranking function $r: Q \times \phi(C) \rightarrow \phi(C)$, where $Q$ is the question set and $\phi(C)$ is the set of all permutations of $C$, such that the answer sentences $A$ are on top of the ranking produced by $r$. The reranker $r$ is often implemented as a pointwise network $f(q, c_i)$, e.g., TANDA \cite{Garg_2020}, which learns a correctness score $p_i \in (0, 1)$ for each answer candidate $c_i$ for ranking. Our work focuses on contextual AS2 \cite{lauriola2021answer}, where additional context such as surrounding sentences is considered to better determine the correctness score for an answer candidate.

Our proposed model, called \textit{``{\asot}''}, for contextual AS2 has four main components: i) Encoding, ii) Question-Context Alignment with OT, iii) Answer-Context Dependencies, and iv) Mutual Information Optimization.

\subsection{Encoding}
\label{sec:encoding}
We are given a question $q = [w^q_1, w^q_2, \ldots, w^q_{T_q}]$ with $T_q$ words and a set of $N$ answer candidates $C = \{c_1, c_2, \ldots, c_N\}$ (retrieved by a search engine), where each candidate is a sentence $c_i = [w^c_1, w^c_2, \ldots, w^c_{T_c}]$ with $T_c$ words. Following the previous work \cite{lauriola2021answer}, we consider previous and next sentences $s_{prev}$, $s_{next}$ as additional context for each answer candidate $c \in C$
\footnote{We employ padding sentences if any context sentence is missing.}.
The input for our model is then formed by concatenating the question, the answer candidate, and the context sentences to obtain a single input sequence: $[q; c; s_{prev}; s_{next}]$. The resulting sequence is fed into a pre-trained language model (PLM), e.g., RoBERTa \cite{liu2019roberta}, to obtain contextualized embeddings for the words. In addition, we employ different segment embeddings for the words in the question, the answer candidate, and the context sentences. These segment embeddings, which are randomly initialized and learnable during training, are added to initial embeddings for the words in the first layer of the PLM. For convenience, let $[\textbf{w}^q_1, \textbf{w}^q_2, \ldots,  \textbf{w}^q_{T_q}]$ and $[\textbf{w}^c_1, \textbf{w}^c_2, \ldots, \textbf{w}^c_{T_c}]$ be the sequences of word representations obtained from the last layer of the PLM for the question $q$ and the answer candidate $c \in C$ respectively.

\subsection{Question-Context Alignment with OT}
\label{sec:alignment}

Optimal Transport (OT) \cite{monge1781memoire,cuturi2013sinkhorn} is an established method to move probability from one distribution to another by finding an alignment between two point sets. In a discrete setting, we are given two probability distributions $p_{X}$ and $p_{Y}$ over two point sets $X = \{ {x_i} \}_{i=1}^n$ and $Y = \{ {y_j} \}_{j=1}^m$ respectively ( $\sum_i p_{x_i} =1$ and $\sum_j p_{y_j} = 1$). A function $D(x, y): X \times Y \rightarrow \mathbb{R}^+$ is also provided to measure the distance between two points $x$ and $y$. OT aims to find a mapping that moves probability mass from the points $\{ {x_i} \}_{i=1}^n$ to the points $\{ {y_j} \}_{j=1}^m$ such that the total cost of transporting the probability mass between the two point sets is minimized. Formally, the goal of OT is to find the transportation matrix $\pi_{XY} \in {\mathbb{R}^+}^{n \times m}$ that minimizes the following transportation cost: $d_{XY} = \sum_{\substack{1 \leq i \leq n \\1 \leq j \leq m}} D({x_i}, {y_j}) {\pi_{XY}}_{ij}$ such that $\pi_{XY}\mathbf{1}_m = p_{X}$ and $\pi_{XY}^T\mathbf{1}_n = p_{Y}$. The transportation matrix $\pi_{XY}$ represents the optimal alignment between the point sets $X$ and $Y$, where the $i$-th row in the matrix provides the optimal alignment from a point $x_i \in X $ to each point $y_j \in Y$. 

In our question-context alignment problem, we consider the question $q$ and a candidate/context sentence $c$ as two point sets: $\{ {w^q_i} \}_{i=1}^{T_q}$ and $\{ {w^c_i} \}_{i=1}^{T_c}$ respectively (each word is a point)\footnote{We exclude stopwords and punctuations from the two point sets before performing the alignment.}. To obtain the probability distributions for the sets, we propose to measure the frequencies of the words and perform a sum normalization. In particular, the probability distribution for the question is computed as follows: $p_{w^q_i} = \frac{freq(w^q_i)}{\sum^{T_q}_{i'=1} freq(w^q_{i'})}$, where $freq(w^q_i)$ is the number of questions that the word $w^q_i$ appears in training data. Next, to estimate the distance between two words (points) $w^q_i \in q$ and $w^c_j \in c$, we measure their semantic divergence by computing the Euclidean distance of their contextualized representations obtained from the PLM: $D(w^q_i, w^c_j) = || \textbf{w}^q_i - \textbf{w}^c_j||$. The optimal transportation matrix $\pi_{XY}$, i.e., $\pi_{qc}$ for the question $q$ and the sentence $c$ can then be solved efficiently using the Sinkhorn-Knopp algorithm \cite{sinkhorn1967concerning,cuturi2013sinkhorn}. Finally, we obtain the relevant context $r_{c}$ for the sentence $c$ as: $r_{c} = \bigcup^{T_q}_{i=1} \{w^c_j| j = \textrm{argmax}_{1\leq j' \leq T_c} {\pi_{qc}}_{ij'}\}$. In the end, we compute the representation for the sentence $c$ as the average sum over the word representations for the relevant context:
\begin{equation}
    \label{eq:candidate-representation}
    \textbf{r}_{c} = \frac{1}{|r_{c}|} \sum_{j | w^c_j \in r_{c}} \textbf{w}^c_j
\end{equation}

\subsection{Answer-Context Dependencies}
\label{sec:dependencies}

For convenience, let $[\textbf{r}_1, \textbf{r}_2, \textbf{r}_3]$ be the representations obtained from Equation (\ref{eq:candidate-representation}) for the answer candidate $s_1 \equiv c$, the previous sentence $s_2 \equiv s_{prev}$, and the next sentence $s_3 \equiv s_{next}$. To learn the dependencies among the sentences, we consider each sentence as a node in a fully-connected graph $G=(V, E)$, where $V = \{s_i\}$ $(1\leq i \leq 3)$ is the node set and $E = \{(s_i, s_j)\}$ ($1 \leq i, j \leq 3$) is the edge set. Our goal is to learn a weight $\alpha_{ij} \in (0, 1)$ for each edge $(s_i, s_j)$ to represent the dependence of $s_i$ on $s_j$. To this end, we propose to employ their semantic representations $\textbf{r}_i$, $\textbf{r}_j$, and transportation costs to the question $d_{qs_i}$, $d_{qs_j}$ to measure the dependency weight $\alpha_{ij}$ between the sentences $s_i$ and $s_j$. Particularly, we first compute the score: $u_{ij} = FFN_{DEP}([\textbf{r}_i \odot \textbf{r}_j; d_{qs_i}; d_{qs_j}])$, where $\odot$ is the element-wise product, $[;]$ represents the concatenation operation, and $FFN_{DEP}$ is a feed-forward network. Afterwards, the weight $\alpha_{ij}$ for the edge $(s_i, s_j)$ is obtained via a softmax function: $\alpha_{ij} = \frac{\textrm{exp}(u_{ij})}{\sum^K_{j'=1} \textrm{exp}(u_{ij'})}$ The induced weights $\{\alpha_{ij}\}$ are then used to enhance the representations for the sentences via $L$ layers of a Graph Convolutional Network (GCN) \cite{kipf:17}:

\begin{equation}
    \textbf{h}^l_i = \textrm{ReLU}(\sum^K_{j=1} \alpha_{ij}\textbf{W}^l\textbf{h}^{l-1}_j + \textbf{b}^l)
\end{equation}
where $\textbf{W}^l$, $\textbf{b}^l$ are learnable weight matrix and bias for the layer $l$ of the GCN ($1 \leq l \leq L$), and $\textbf{h}^0_i \equiv \textbf{r}_i$ is the input representation for the sentence $s_i$. The output vectors $\textbf{h}^{L}_i \equiv \textbf{h}_i$ at the last layer of the GCN serve as the final representations for the sentences $s_i$. The representation $\textbf{h}_1$ for the answer candidate $s_1 \equiv c$ is finally sent to a feed-forward network with a sigmoid output function to estimate the correctness score $p_c \in (0, 1)$ for the answer candidate $c$: $p_c = FFN_{AS2}(\textbf{h}_1)$. For training, we minimize the binary cross-entropy loss $L_{AS2}$ with the correctness scores $p_c$.

\subsection{Mutual Information Optimization}
\label{sec:mi-optimization}

In information theory, MI is defined as the KL divergence between the joint distribution and the product of the marginal distributions of two random variables. As a result, two random variables would be more dependent if they have large mutual information. In our case, some of the context sentences might also be correct/incorrect answers for the question. Therefore, we expect answer sentence pairs to share more semantic information and (answer, non-answer) sentence pairs to share less semantic information relevant to the question. This can be done by considering sentence representation vectors $\textbf{h}_i$ as random variables and maximizing/minimizing the mutual information between the variables, respectively. However, the sentence vectors $\textbf{h}_i$ are very high dimensional variables, making the exact calculation of the MI between the vectors impossibly expensive. To overcome this, we followed the mutual information neural estimation (MINE) method \cite{belghazi2018mutual,hjelm2018learning} to estimate and optimize the lower bound of the MI between the variables via the binary cross entropy of a variable discriminator $U$, which is a feed-forward network with a sigmoid output function:
\begin{multline}
\label{eq:mi-loss}
    L_{MI} = - \sum_{(i, j) \in I_+} \textrm{log}(U([\textbf{h}_i; \textbf{h}_j]) \\- \sum_{(i', j') \in I_-} \textrm{log} (1 - U([\textbf{h}_{i'}; \textbf{h}_{j'}])
\end{multline}
where $I_+ = \{(i, j) | s_i, s_j \in A\} (1\leq i, j \leq 3)$ is the index set for answer sentence pairs, and $I_- = \{(i', j') | s_{i'} \in A, s_{j'} \notin A\} (1\leq i', j' \leq 3)$ is the index set for (answer, non-answer) sentence pairs among the three sentences.

\subsection{Training and Inference}
All the components in our proposed model are jointly trained via minimizing the loss function:
\begin{equation}
    L = L_{AS2} + \gamma L_{MI}
\end{equation}
where $\gamma$ is a hyper-parameter to balance the contributions of each component to the training of the model. Following the previous work \cite{Garg_2020,lauriola2021answer}, we consider all answer candidates for each question for training and inference.

\begin{table}[ht]
\centering
{
\begin{tabular}{|l|r|r|r|r|}
\hline
\textbf{Datasets} & \textbf{Train} & \textbf{Dev} & \textbf{Test} & \textbf{\#A/Q} \\ \hline
WikiQA            & 2,118            & 126          & 243           & 1.18           \\ \hline
WDRASS            & 53,419         & 5,416        & 5,395         & 4.96           \\ \hline
\end{tabular}
}
\caption{The number of questions in training, development, and test data of the two datasets. The last column presents the average number of answers to a question.}
\label{tab:datasets}
\end{table}

\section{Results}

\subsection{Experimental Setup}
\label{sec:experimental-setup}

\noindent \textbf{Datasets} Following the previous works \cite{Garg_2020,lauriola2021answer}, we use the same train/dev/test splits for the standard AS2 dataset, i.e., \textit{WikiQA} \cite{yang-etal-2015-wikiqa}. In addition, we also experimented with a large-scale AS2 dataset called \textit{WDRASS} \cite{zhang2022situ} to investigate the models' performance further.

\begin{itemize}
    \item \textbf{WikiQA} is a QA dataset created by \cite{yang-etal-2015-wikiqa}. The dataset contains questions and answer candidates, manually annotated on Bing query logs over Wikipedia. Following the previous works, we conduct experiments with the \textit{clean} version of the dataset, and combine the development and test sets to obtain a larger and more reliable set for model comparison.
    \item \textbf{WDRASS} is recently created by \cite{zhang2022situ}. WDRASS is a large-scale dataset focusing on non-factoid questions requiring entire sentences to answer.
\end{itemize}

Statistics for all the datasets are shown in Table 2. 

\noindent \textbf{Hyper-parameters and Tools} Following the previous work \cite{lauriola2021answer}, we use a small portion of the WikiQA training data to tune hyper-parameters for our model and select the best hyper-parameters for all the datasets. We employ Adam optimizer to train the model with a learning rate of $1e-5$ and a batch size of $64$. We set $400$ for the hidden vector sizes for all the feed-forward networks, $L=2$ for the number of the GCN layers, and $0.3$ for the trade-off weights $\gamma$. To implement the models, we use Pytorch version 1.7.1 
and Huggingface Transformers version 3.5.1.
We use the NLTK library version 3.5 \cite{bird2009natural} to preprocess the data and remove stopwords. The model performance is obtained over three runs with different random seeds.

\noindent \textbf{Evaluation Metrics} Following the previous works, we measure the model performance using the entire set of answer candidate sentences for each question, using the three metrics: Precision-at-1 (P@1), Mean Average Precision (MAP), and Mean Reciprocal Rank (MRR) scores.

\subsection{Performance Comparison}
We compare our proposed model with TANDA \cite{Garg_2020} and LOCT \cite{lauriola2021answer}, which are the current state-of-the-art models for AS2. Table \ref{tab:main-results} shows the perforformance comparison between the models on two settings: i) using a non-finetuned RoBERTa base encoder, and ii) using a finetuned RoBERTa base encoder. The non-finetuned RoBERTa Base is obtained from \cite{liu2019roberta} while the other is produced by finetuning TANDA on the ASNQ dataset \cite{Garg_2020}. As can be seen from the table, all the models benefit from using the finetuned RoBERTa Base encoder. Across the two settings, our model outperforms the previous models by large margins, demonstrating its effectiveness for AS2.
\begin{table}[]
\small
\centering
\begin{tabular}{|l|ll|ll|}
\hline
\multirow{2}{*}{Models} & \multicolumn{2}{c|}{\begin{tabular}[c]{@{}c@{}}Non-finetuned\\ RoBERTa Base\end{tabular}} & \multicolumn{2}{c|}{\begin{tabular}[c]{@{}c@{}}ASNQ-finetuned\\ RoBERTa Base\end{tabular}} \\ \cline{2-5} 
                        & \multicolumn{1}{c|}{P@1}                      & \multicolumn{1}{c|}{MAP}                  & \multicolumn{1}{c|}{P@1}                      & \multicolumn{1}{c|}{MAP}                   \\ \hline
TANDA                   & \multicolumn{1}{l|}{63.24*}                   & 75.00*                                    & \multicolumn{1}{l|}{78.67*}                   & 86.74*                                     \\ \hline
LOCT           & \multicolumn{1}{l|}{68.09*}                   & 79.00*                                    & \multicolumn{1}{l|}{81.31*}                   & 88.00*                                     \\ \hline
\textbf{{\asot}}     & \multicolumn{1}{l|}{\textbf{74.16}}                    & \textbf{83.29}                                     & \multicolumn{1}{l|}{\textbf{83.77}}                    & \textbf{89.28}                                      \\ \hline
\end{tabular}
\caption{Performance comparison of the models on combination of developement and test sets of WikiQA. * indicates the performance officially reported by the previous work \cite{lauriola2021answer}.}
\label{tab:main-results}
\end{table}

\begin{table}[]
\small
\centering
\begin{tabular}{|l|lll|}
\hline
\multicolumn{1}{|c|}{\multirow{2}{*}{Models}} & \multicolumn{3}{c|}{\begin{tabular}[c]{@{}c@{}}ASNQ-finetuned\\ RoBERTa Base\end{tabular}} \\ \cline{2-4} 
\multicolumn{1}{|c|}{}                        & \multicolumn{1}{c|}{P@1}      & \multicolumn{1}{c|}{MAP}     & \multicolumn{1}{c|}{MRR}    \\ \hline
TANDA                                         & \multicolumn{1}{l|}{54.6}     & \multicolumn{1}{l|}{63.5}    & 64.3                        \\ \hline
\textbf{{\asot}}                           & \multicolumn{1}{l|}{\textbf{55.9}}     & \multicolumn{1}{l|}{61.8}    & \textbf{69.7}                        \\ \hline
\textbf{{\asot}} (joint)                               & \multicolumn{1}{l|}{\textbf{55.9}}     & \multicolumn{1}{l|}{\textbf{64.2}}    & \textbf{65.0}                        \\ \hline
\end{tabular}
\caption{Performance comparison on WDRASS test set.}
\label{tab:wdrass-results}
\end{table}

Table 4 shows the performance of our proposed model compared to TANDA on the WDRASS test set. {\asot} significantly improves the performance for P@1 and MRR, however, decreases the performance for MAP. We attribute this to the fact that our model ranks the answer candidates individually. To deal with this, we explore another use case of our model where it can produce a joint reranking for multiple answer candidates ranked by TANDA, leading to significantly better performance for all the three metrics.


\section{Conclusions and Future Work}

In this work, we propose {\asot}, a novel LLM-based model that (i) efficiently learns the answer-context dependencies to improve representation learning for AS2 by (ii) leveraging relevant context in answer/context sentences captured via question-context alignments using Optimal Transport.
Experimental results demonstrate the efficacy of our proposed model, resulting in significant improvements and new state-of-the-art performance across several widely-used AS2 benchmark WikiQA \cite{yang-etal-2015-wikiqa} and a recent large-scale AS2 dataset (WDRASS) \cite{zhang2022situ}.

Naturally, our proposed model {\asot} can approximate human judgement for the correctness of an answer for a given question. As a result, the model can be used to provide signals for assessing quality of answers produced by open-domain question answering (ODQA) systems. As such, we are exploring AS2 datasets and architecture designs for building better reward models for reinforcement learning from human feedback \cite{nakano2021webgpt}, which has shown impressive improvements for ODQA via training LLMs to generate answers for open-domain questions.

\section{Acknowledgments}
We would like to thank Zeyu Zhang for sharing the WDRASS dataset \cite{zhang2022wdrass} and his helpful comments on the paper regarding the experiment setup and data preparation.

\bibliography{mybib, anthology-filter}
\bibliographystyle{IEEEtran}

\end{document}